# Coupled Analysis Dictionary Learning to inductively learn inversion: Application to real-time reconstruction of Biomedical signals

Kavya Gupta
Embedded Systems and Robotics,
TCS Research and Innovation, India
gupta.kavya@tcs.com

Brojeshwar Bhowmick
Embedded Systems and Robotics,
TCS Research and Innovation, India
b.bhowmick@tcs.com

Angshul Majumdar
Indraprastha Institute of Information
Technology Delhi, India
angshul@iiitd.ac.in

*Abstract*— **This work addresses the problem of reconstructing biomedical signals from their lower dimensional projections. Traditionally Compressed Sensing (CS) based techniques have been employed for this task. These are transductive inversion processes; the problem with these approaches is that the inversion is time-consuming and hence not suitable for real-time applications. With the recent advent of deep learning, Stacked Sparse Denoising Autoencoder (SSDAE) has been used for learning inversion in an inductive setup. The training period for inductive learning is large but is very fast during application – capable of real-time speed. This work proposes a new approach for inductive learning of the inversion process. It is based on Coupled Analysis Dictionary Learning. Results on Biomedical signal reconstruction show that our proposed approach is very fast and yields result far better than CS and SSDAE.**

## I. INTRODUCTION

This work is particularly pertinent to the topic of telemonitoring of biomedical signals via Wireless Body Area Network (WBAN). In a WBAN the signals are collected by the sensor nodes and are transmitted to a remote base station for further processing and analysis. The sensor nodes have limited computational power and power backup. Therefore one of the most pressing problems in such scenarios is how to acquire and transmit the signals in an energy and computationally efficient manner.

There are three power sinks at the sensor nodes – acquisition, processing, and transmission. The last one consumes the most power followed by acquisition; the energy requirement for processing is negligible as compared to the other two. Ideally one would like to acquire the full signal, compress it using transform coding techniques and transmit the compressed version. However, transform coding is computationally expensive and cannot be implemented on the hardware at the sensor nodes.

This leads to the idea of compressing the acquired signal by random projections – this is very cheap to compute. The lower dimensional projections are then transmitted. There is enough computational power at the base station to carry out Compressed Sensing (CS) based reconstruction of the signals [1], [2], [3].

CS based techniques and its variants (dictionary learning / blind compressed sensing) [4] are slow; they require solving complex optimization problems. Hence the compressed sensing based reconstruction paradigm is not suitable for real-time operations. For example, in epileptic seizure prediction [5] from EEG or coronary ischemia prediction [6] from ECG, one does not have the luxury of losing time in reconstructing the signal. For such critical care health monitoring scenarios, when the time is of essence, one needs to reconstruct the signal in real-time.

CS involves no learning. Dictionary learning based techniques are inductive in nature, i.e. they learn from the signal that they have to reconstruct. With the advent of deep learning new inductive techniques for solving inverse problems have been proposed. The most generic ones are based on Stacked Sparse Denoising Autoencoder (SSDAE) [7], [8], [9], [10]. There are a few others that are specific for images [11], [12], [13] based on the convolutional neural networks.

These inductive techniques apply the transpose of the projection operator on the lower dimensional measurements to get a noisy signal. This is applied at the input and the clean version of it is applied at the output during the learning process. The deep neural network 'learns' the inversion process. The problem with all such approaches is that they can only work in the signal domain and not from the lower dimensional measurements. In this work, we propose a technique where the inversion will be learned directly from the measurement domain.

Our work is based on the coupled representation learning approach. The main idea in coupled representation learning is to learn a basis (and corresponding coefficients) for the two domains - source and target, such that the coefficients from one domain can be linearly mapped to the other. We learn a representation in the source domain, which in our case constitutes the lower dimensional measurements; the target domain is the signal – a representation of the signal is learned as well. There is a learned coupling map from the source (measurement domain) to the target (signal domain) representation. Coupled Autoencoders (CAE) [14], [15], [16] and Coupled Dictionary Learning (SDL) [17], [18] have been proposed before. In this work, we introduce Coupled Analysis Dictionary learning or Coupled transform learning.

## II. LITERATURE REVIEW

### A. Compressed Sensing Based Reconstruction

The biomedical signal '$x$' is collected at the sensor node. For energy efficient transmission, it needs to be compressed. Compressed Sensing (CS) compresses the signal by projecting it onto a random matrix. This is expressed as,

$$y = Ax \quad (1)$$

Here $x$ is the signal; $A$ is the compression matrix and $y$ the compressed measurement.

This compressed measurement is transmitted to a remote base station where the signal is reconstructed by exploiting its sparsity in some transform domain like DCT or wavelet. This is expressed as,

$$\min_{\alpha} \|\alpha\|_1 \text{ such that } \|y - AS^T\alpha\|_2^2 \le \varepsilon \quad (2)$$

Here $\alpha$ is the sparse transform coefficient in domain $S$ for the signal $x$, $\varepsilon$ is a parameter that controls the data fidelity. It is assumed that the sparsifying transform is either orthogonal or tight-frame.

This formulation (or slight variants) has been used in [1], [2], [3]. In the Blind Compressed Sensing (BCS) approach instead of assuming that the signal is sparse in some transform $S$, the basis is learned from the data. This is similar to dictionary learning based reconstruction techniques. However, since biomedical signals are small, patch-based techniques used in dictionary learning does not yield good results. However, when multi-channel data is considered, BCS can be used [4]. But, BCS has restricted application. It cannot be used on single channel data such as Photoplethysmogram (PPG). It cannot be used when the number of channels is few for example in ECG or in non-medical grade EEG.

### B. Stacked Denoising Autoencoder

An autoencoder is a self-supervised neural network, i.e. the input and the output are same. There is an encoder $W$ that projects the input to a hidden representation and a decoder ($W'$) that reverse maps the representation to the output (= input). Mathematically this is expressed as,

$$X = W'\varphi(WX) \quad (3)$$

Here $\varphi$ is a non-linear activation function.

Given the training samples ($X$), the encoder and the decoder are estimated by minimizing the Euclidean cost function.

$$\min_{W,W'} \|X - W'\varphi(Wx)\|_F^2 \quad (4)$$

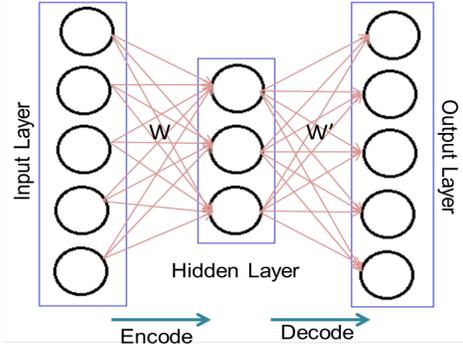

Fig.1. Basic Autoencoder Structure.

The shallow (single layer) autoencoder is shown in Fig.1. Deeper architectures can be learned by nesting autoencoders inside each other. There are multiple encoders followed by an equal number of decoders. The learning is expressed as,

$$\arg\min_{W_1......W_L, W_1'....W_L'} \|X - g \circ f(X)\|_F^2 \quad (5)$$

where $g = W_1'\phi(W_2'....W_L'(f(X)))$ and $f = \phi(W_L\phi(W_{L-1}......\phi(W_1 X)))$.

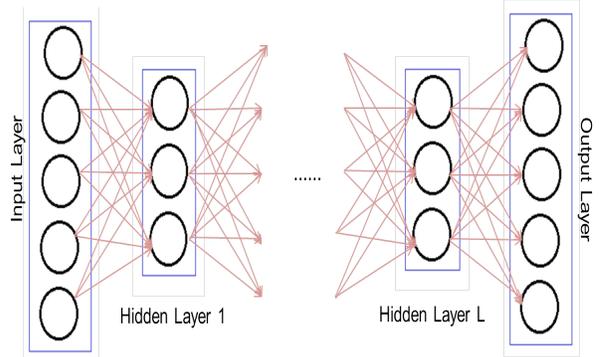

Fig.2. Stacking of Autoencoders

Traditionally autoencoders have been used to pre-train deep neural networks [19]. However that is not the objective of this work, so we will not pursue discussion on that topic. Studies in the recent past [8], [9], [10] have shown that autoencoders can be used for 'learning to solve' inverse problems; especially simple inverse problems like denoising and deblurring. During the training phase, the corrupt signal (noisy/blurry) is input to the autoencoder and the corresponding clean signal is at the output. From a large volume of data, the stacked autoencoder learns the inversion operation. During operation the corrupt signal is input, the stacked autoencoder is expected to clean it.

For images, this idea has been extended to reconstruction. A corrupted version from compressive measurements can be obtained via application of the transpose of the measurement operator, i.e. $\hat{x} = A^T y$. This corrupt version is used at the input and the clean one at the output during training. However, note that such studies have only worked on images using a CNN framework [11], [12], [13]. But in principle can be used for other signals as well.

C. Transform Learning

Dictionary learning as shown in Fig.3 is a well-studied topic, but transform learning in Fig.4 is relatively new. Hence we discuss it briefly for ease of the reader.

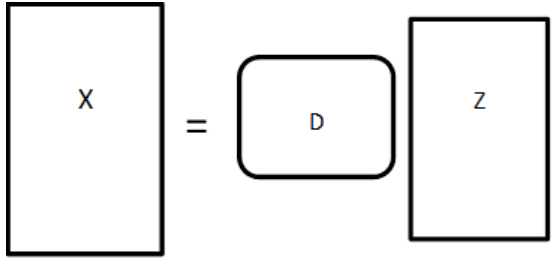

Fig.3. Dictionary Learning

Analysis Dictionary learning or transform learning analyses the data by learning a basis to produce coefficients. Mathematically this is expressed as,

$$TX = Z \quad (6)$$

Here $T$ is the transform/analysis basis, $X$ is the data and $Z$ is the corresponding coefficients.

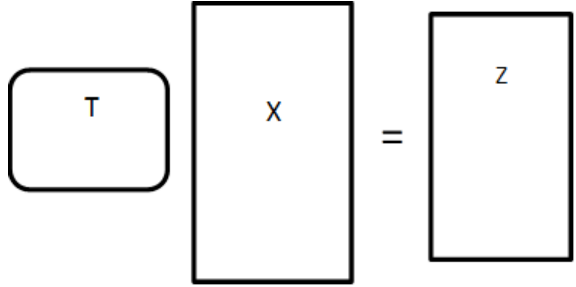

Fig.4. Analysis Dictionary Learning

The following analysis dictionary learning formulation was proposed in [20], [21] as,

$$\min_{T,Z} \|TX - Z\|_F^2 + \lambda \left( \|T\|_F^2 - \log \det T \right) \quad (7)$$

The factor $(-\log \det T)$ imposes a full rank constraint on the learned transform; this prevents the degenerate solution ($T=0$, $Z=0$). The additional penalty $\|T\|_F^2$ is to balance scale; without this $(-\log \det T)$ can keep on increasing, producing degenerate results in the other extreme.

In [20], [21], an alternating minimization approach was proposed to solve the said problem. Z and T updates are given by,

$$\hat{Z} \leftarrow \min_Z \|TX - Z\|_F^2 \quad (8a)$$

$$\hat{T} \leftarrow \min_T \|TX - Z\|_F^2 + \lambda \left( \|T\|_F^2 - \log \det T \right) \quad (8b)$$

Updating the coefficients Z (8a) is straightforward - simple least squares, having an analytic solution in the form of pseudoinverse. For updating the transform (8b), the gradients for different terms in (8b) are easy to compute. Ignoring the constants this is given by,

$$\nabla \|TX - Z\|_F^2 = X^T (TX - Z)$$

$$\nabla \|T\|_F^2 = T$$

$$\nabla \log \det T = T^{-T}$$

In the initial paper on transform learning [20], a non-linear conjugate gradient based technique was proposed to solve the transform update (8b). In the more refined version [21], with some linear algebraic tricks, they were able to show that a closed form update exists for the transform using Cholesky decomposition and Singular Value Decomposition.

$$XX^T + \lambda I = LL^T \quad (9a)$$

$$L^{-1} XZ^T = USV^T \quad (9b)$$

$$T = 0.5V \left( S + (S^2 + 2\lambda I)^{1/2} \right) U^T L^{-1} \quad (9c)$$

The proof for convergence of such an update algorithm can be found in [22].

III. COUPLED ANALYSIS DICTIONARY LEARNING

In Coupled Analysis Dictionary learning there is a measurement domain ($M$), and there is a signal domain ($S$). Coupled learning learn two analysis basis / transforms $T_M$ and $T_S$ (one for each domain) and their corresponding features $Z_M$ and $Z_S$ so that the features from one of the domains can be linearly mapped ($A$) into the other.

Mathematically our formulation is expressed as,

$$\min_{T_M, T_S, Z_M, Z_S, A} \|T_M Y - Z_M\|_F^2 + \|T_S X - Z_S\|_F^2$$
$$+ \lambda \left( \|T_M\|_F^2 + \|T_S\|_F^2 - \log \det T_M - \log \det T_S \right)$$
$$+ \mu \|Z_S - AZ_M\|_F^2$$
$$(10)$$

Here $Y$ consists of the training measurements stacked as columns, $X$ the corresponding signals. $Z_M$ and $Z_S$ are the coefficients in the measurement and signal domains respectively.

The alternating minimization approach is used for solving (10). It can be segregated into the following sub-problems.

$$P1: \min_{T_M} \|T_M Y - Z_M\|_F^2 + \lambda \left( \|T_M\|_F^2 - \log \det T_M \right)$$

$$P2: \min_{T_S} \|T_S X - Z_S\|_F^2 + \lambda \left( \|T_S\|_F^2 - \log \det T_S \right)$$

$$P3: \min_{Z_M} \|T_M Y - Z_M\|_F^2 + \mu \|Z_S - A Z_M\|_F^2$$

$$\Rightarrow \min_{Z_M} \left\| \begin{pmatrix} T_M Y \\ \sqrt{\mu} Z_S \end{pmatrix} - \begin{pmatrix} I \\ \sqrt{\mu} A \end{pmatrix} Z_M \right\|_F^2$$

$$P4: \min_{Z_S} \|T_S X - Z_S\|_F^2 + \mu \|Z_S - A Z_M\|_F^2$$

$$\Rightarrow \min_{Z_S} \left\| \begin{pmatrix} T_S X \\ \sqrt{\mu} A Z_M \end{pmatrix} - \begin{pmatrix} I \\ \sqrt{\mu} I \end{pmatrix} Z_S \right\|_F^2$$

$$P5: \min_A \|Z_S - A Z_M\|_F^2$$

Sub-problems P1 and P2 are standard transform updates. We already know how to update them (9). Sub-problem P3, P4, and P5 are simple least square problems and this concludes the training phase.

During operation, we will have the compressed version $y$ in the measurement/source domain. From this, we will find the corresponding coefficients by,

$$z_M = T_M y \qquad (11)$$

From the features of the measurement domain, the signal domain features are generated by $\hat{z}_S = A z_M$. From these features, the corresponding target domain signal ($x$) is synthesized by solving,

$$T_S x = \hat{z}_S \qquad (12)$$

This has an analytic solution in the form of the pseudo-inverse.

## IV. EXPERIMENTAL RESULTS

In this work, experiments were conducted on UCI Cuff-less Blood Pressure Estimation dataset [23]. The dataset consists of Photoplethysmograph (PPG) from the fingertip, arterial blood pressure (ABP) and electrocardiogram (ECG) from channel II. All of them have been sampled at 125 Hz.

For the experiments, we divide the signals into chunks of 512 samples which served as ground truth. The training data consisted of 75374 such samples and the testing data consisted of 13448 samples. For emulating the wireless telemonitoring scenario, these signals were projected onto a lower dimension space by Bernoulli projection matrix with 25% and 50% undersampling. These lower dimensional measurements served as the measurement domain (M) and the corresponding ground truth as the signal domain(S). At the testing end, we reconstruct the signal from its lower dimensional representation.

There are two parameters to be tuned to the proposed method – λ and μ. The final values used here are λ=0.1 and μ=1. The transforms for the signal and the measurement domain ($T_S$ and $T_M$) are perfectly determined. The iterations were run until the objective function converged to local minima. By convergence, we mean that the value of the objective function does not change much in successive iterations.

We compare the proposed method with two other techniques. The first one is Compressive Sensing (CS) [2] and other is Stacked Sparse Denoising Autoencoder (SSDAE); these have been used in the past for reconstruction [24], [25].

We tried our best to optimize SPGL1 and SSDAE on these signals for a fair comparison to the proposed approach. For PPG and ABP a 3 layered autoencoders is used 256-128-64 which are learned greedily with an Absolute Criterion as a Loss function. For ECG only one layer of autoencoders was used with an Absolute Criterion for loss; this is because increasing the number of layers was detrimental to its performance. Network for SSDAE was built in Torch [26].

TABLE I: RESULTS ON PPG

| Undersampling Ratio | Error | SPGL1 | SSDA | Proposed |
|---|---|---|---|---|
| 0.25 | Mean, ± std | 0.16, ± .05 | 0.07, ±.021 | **0.01, ± .01** |
| | Max | 0.35 | 0.49 | **0.16** |
| | Min | 0.03 | 0.03 | **0.00** |
| 0.50 | Mean, ± std | 0.06, ± .02 | 0.05, ±.02 | **0.00, ± .00** |
| | Max | 0.16 | 0.49 | **0.07** |
| | Min | 0.01 | 0.02 | **0.00** |

TABLE II: RESULTS ON ABP

| Undersampling Ratio | Error | SPGL1 | SSDAE | Proposed |
|---|---|---|---|---|
| 0.25 | Mean, ± std | 0.16, ± .04 | 0.06, ± .01 | **0.01, ±.01** |
| | Max | 0.34 | 0.19 | **0.10** |
| | Min | 0.07 | 0.03 | **0.00** |
| 0.50 | Mean, ± std | 0.06, ±.01 | 0.05, ±.01 | **0.00, ±.00** |
| | Max | 0.14 | 0.17 | **0.05** |
| | Min | 0.03 | 0.02 | **0.00** |

TABLE III: RESULTS ON ECG

| Undersampling Ratio | Error | SPGL1 | SSDAE | Proposed |
|---|---|---|---|---|
| 0.25 | Mean, ± std | 0.17, ±.02 | 0.43, ±.02 | **0.05, ±.03** |
| | Max | 0.44 | 0.68 | **0.38** |
| | Min | 0.10 | 0.40 | **0.00** |
| 0.50 | Mean, ± std | 0.07, ±.01 | 0.41, ±.02 | **0.03, ±.01** |
| | Max | **0.21** | 0.63 | 0.25 |
| | Min | 0.05 | 0.38 | **0.00** |

All the experiments were run on an Intel Xeon 3.50GHz × 8 Processor, having 64 GB RAM. For SSDAE Nvidia Quadro 4000 GPU was used. Table 4 shows the timings in seconds. The training time is for all the samples in the train sets and test time is shown per test sample to show the applicability in the real-time scenario. Since all the datasets are of the same size, we only show the results for PPG; the timings for the other two types of signal are almost the same (SSDAE for ECG has slightly less training time since single layer was used). We can see that the reconstruction time for SPGL1 samples is higher than the permissible amount for real-time reconstruction. Our proposed method and autoencoder based techniques are very fast – capable of real-time reconstruction. Ours is orders of magnitude faster compared to the autoencoder since we need fewer matrix-vector products.

TABLE IV: TRAINING AND TESTING TIMES IN SECONDS

| Method | Training Time(s) | Testing Time(s)/sample |
|---|---|---|
| SPGL1 | - | 1.5 |
| SSDAE | 18360 | $4.6 \times 10^{-3}$ |
| Proposed | 70 | $1.1 \times 10^{-5}$ |

V. CONCLUSION

This work introduces Coupled transform learning/analysis Dictionary learning. This is a generic formulation for domain adaptation/transfer learning. However, the specific interest of this work was to solve inversion problems. We show that by modeling the measurement as the source domain and the signal as the target domain, we can learn the inversion operation via our proposed formulation. This is an inductive inversion approach. In this work, we propose a technique where the inversion will be learned directly from the measurement domain, which was missing in the previous inductive approaches. Comparison with Stacked Sparse Denoising Autoencoders and Compressed Sensing show that our method is considerably superior in terms of reconstruction accuracy and is very fast – capable of real-time application.

Transfer learning finds a variety of applications in computer vision and natural language processing. In vision, it has been used for synthesis problems like deblurring and super-resolution and for analysis problems like RGB to NIR matching and photo to sketch matching. In NLP it has been used for multi-lingual document retrieval. It has also been used for multi-modal (text to video) retrieval. In the future, we would like to investigate our proposed approach to such problems.

ACKNOWLEDGMENT

Angshul Majumdar has been partially supported by the 5IOA036 FA23861610004 grant by Air Force Office of Scientific Research (AFOSR), AOARD.

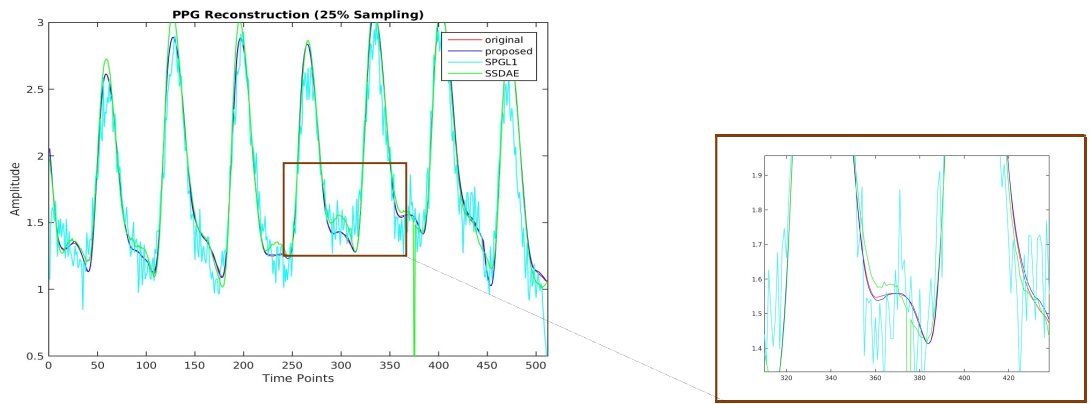

(a)

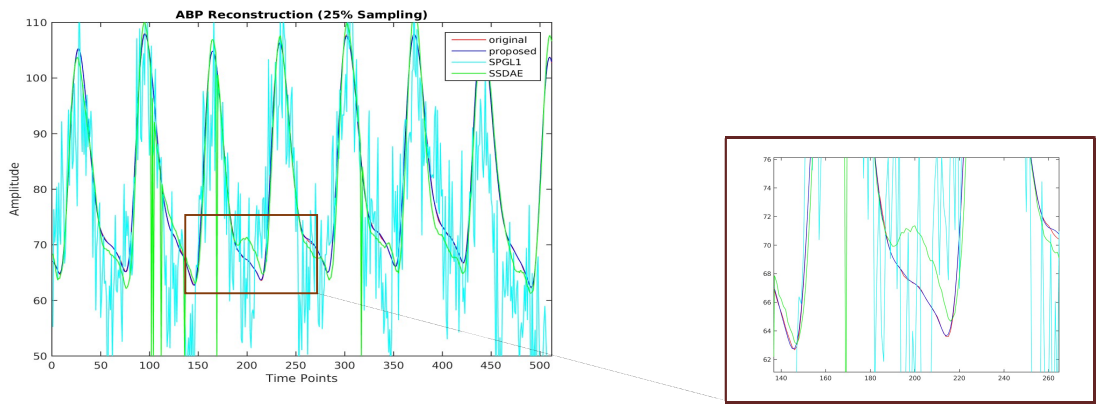

(b)

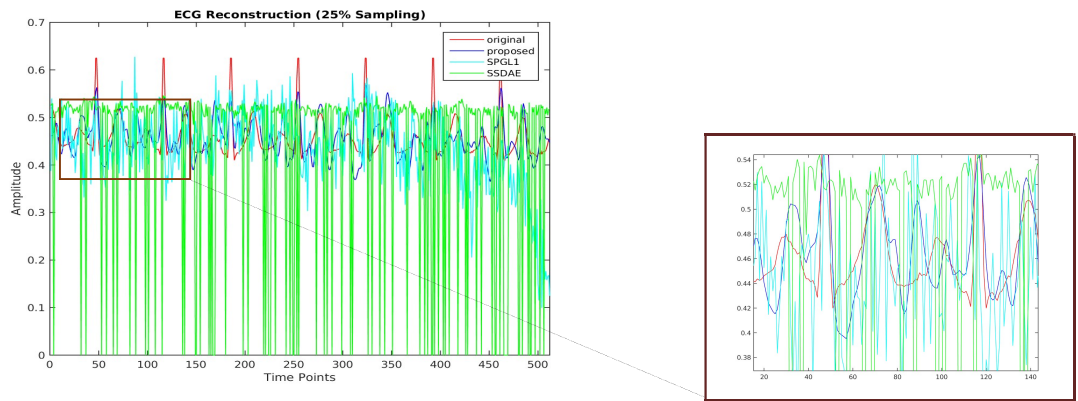

(c)

Fig. 5. Top to Bottom – PPG, ABP, and ECG. Reconstruction at 25% undersampling. The figures to the right show the zoomed inset to better understand the difference between proposed and previous techniques.

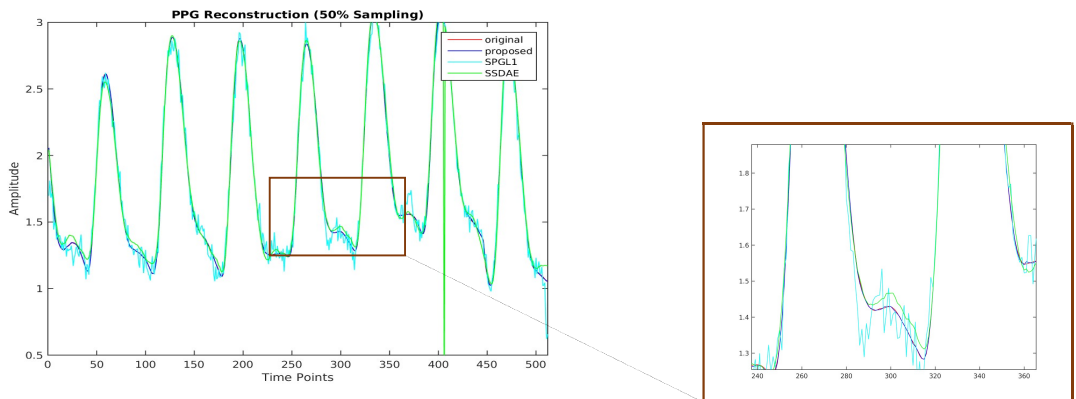

(a)

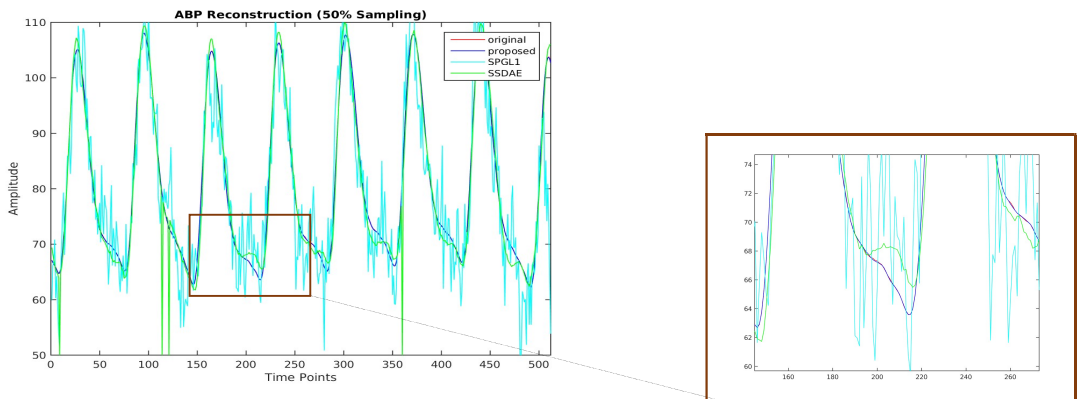

(b)

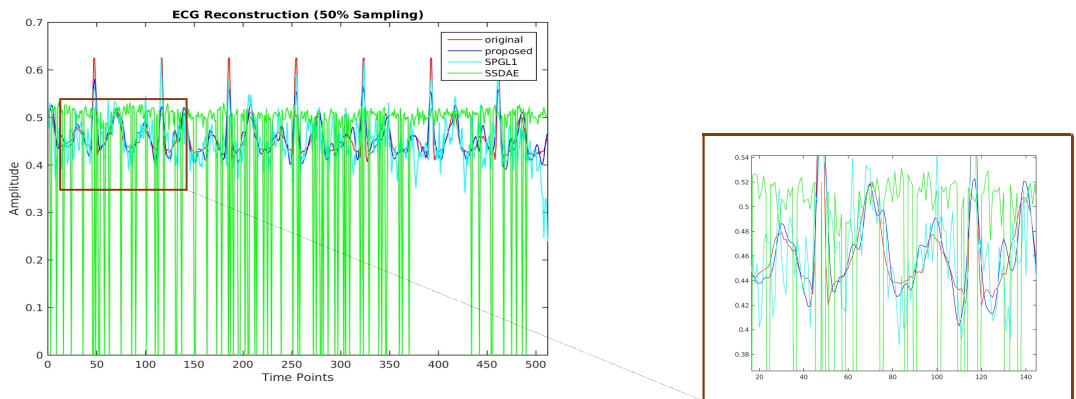

(c)

Fig. 6. Top to Bottom – PPG, ABP, and ECG. Reconstruction at 50% undersampling. The figures to the right show the zoomed inset to better understand the difference between proposed and previous techniques.

# REFERENCES

[1] S. Aviyente, "Compressed Sensing Framework for EEG Compression", IEEE Workshop on Statistical Signal Processing, pp.181,184, 2007.

[2] Z. Zhang, T. P. Jung, S. Makeig and B. D. Rao, "Compressed Sensing of EEG for Wireless Telemonitoring with Low Energy Consumption and Inexpensive Hardware", IEEE Transactions on Biomedical Engineering, Vol. *60*(1), 221-224, 2013.

[3] M. Mohsina and A. Majumdar, "Gabor Based Analysis Prior Formulation For EEG Signal Reconstruction", Biomedical Signal Processing and Control, Vol. 8(6), pp.951-955, 2013.

[4] A. Majumdar and A. Shukla, "Row-sparse Blind Compressed Sensing for Reconstructing Multi-channel EEG signals", Biomedical Signal Processing and Control. Vol. 18, pp.174-178, 2015.

[5] L. Chisci et al., "Real-Time Epileptic Seizure Prediction Using AR Models and Support Vector Machines," in IEEE Transactions on Biomedical Engineering, vol. 57, no. 5, pp. 1124-1132, May 2010.

[6] F. A. Khan, N. A. H. Haldar, A. Ali, M. Iftikhar, T. A. Zia and A. Y. Zomaya, "A Continuous Change Detection Mechanism to Identify Anomalies in ECG Signals for WBAN-Based Healthcare Environments," in IEEE Access, vol. 5, pp. 13531-13544, 2017.

[7] J. Mehta and A. Majumdar, "RODEO: Robust DE-aliasing autoencOder for Real-time Medical Image Reconstruction", Pattern Recognition, Vol. 63, pp. 499-510, 2017.

[8] F. Agostinelli, M. R. Anderson and H. Lee, "Adaptive multi-column deep neural networks with application to robust image denoising", Neural Information Processing Systems, pp. 1493-1501, 2013.

[9] J. Xie, L. Xu and E. Chen, "Image denoising and inpainting with deep neural networks", Neural Information Processing Systems, pp. 341-349, 2012.

[10] H. C. Burger, C. J. Schuler and S. Harmeling, "Image denoising: Can plain neural networks compete with BM3D?" IEEE Conference on Computer Vision and Pattern Recognition, pp. 2392-2399, 2012.

[11] K. Kulkarni, S. Lohit, P. Turaga, R. Kerviche and A. Ashok, "ReconNet: Non-iterative reconstruction of images from compressively sensed random measurements" IEEE Conference on Computer Vision and Pattern Recognition, pp. 449-458, 2016.

[12] K. Xu, and F. Ren, "CSVideoNet: A Recurrent Convolutional Neural Network for Compressive Sensing Video Reconstruction". arXiv preprint arXiv:1612.05203, 2016.

[13] C.A Metzler, A. Mousavi, and R.G Baraniuk, "Learned D-AMP: A Principled CNN-based Compressive Image Recovery Algorithm.", arXiv preprint arXiv:1704.06625, 2017.

[14] K. Zeng, J. Yu, R. Wang, C. Li and D. Tao, "Coupled Deep Autoencoder for Single Image Super-Resolution," in IEEE Transactions on Cybernetics, vol. 47, no. 1, pp. 27-37, Jan. 2017.

[15] B. S. Riggan, C. Reale and N. M. Nasrabadi, "Coupled Auto-Associative Neural Networks for Heterogeneous Face Recognition," in IEEE Access, vol. 3, pp. 1620-1632, 2015.

[16] K. Gupta, B. Bhowmick and A. Majumdar, "Motion Blur Removal via Coupled Autoencoder", IEEE ICIP, 2017.

[17] D. A. Huang and Y. C. F. Wang, "Coupled Dictionary and Feature Space Learning with Applications to Cross-Domain Image Synthesis and Recognition," 2013 IEEE International Conference on Computer Vision, Sydney, NSW, 2013, pp. 2496-2503.

[18] S. Wang, L. Zhang, Y. Liang and Q. Pan, "Semi-coupled dictionary learning with applications to image super-resolution and photo-sketch synthesis," 2012 IEEE Conference on Computer Vision and Pattern Recognition, Providence, RI, 2012, pp. 2216-2223.

[19] P. Vincent, H. Larochelle, I. Lajoie, Y. Bengio and P. A. Manzagol, "Stacked denoising autoencoders: Learning useful representations in a deep network with a local denoising criterion", The Journal of Machine Learning Research, Vol. 11, pp. 3371-3408, 2010.

[20] S. Ravishankar and Y. Bresler, "Learning sparsifying transforms", IEEE Transactions on Signal Processing, Vol. 61 (5), pp. 1072-1086, 2013.

[21] S. Ravishankar, B. Wen and Y. Bresler. "Online sparsifying transform learning-Part I: Algorithms", IEEE Journal of Selected Topics in Signal Processing, Vol. 9 (4), pp. 625-636, 2015.

[22] S. Ravishankar and Y. Bresler, "Online Sparsifying Transform Learning-Part II: Convergence Analysis", IEEE Journal of Selected Topics in Signal Processing, Vol. 9 (4), pp. 637-646, 2015.

[23] https://archive.ics.uci.edu/ml/datasets/Cuff-Less+Blood+Pressure+Estimation#

[24] A. Majumdar, A. Gogna and R. K. Ward, "Semi-supervised Stacked Label Consistent Autoencoder for Reconstruction and Analysis of Biomedical Signals", IEEE Transactions on Biomedical Engineering, Vol. 64 (9), pp. 2196 – 2205, 2017.

[25] A. Majumdar and R. Ward, "Real-time reconstruction of EEG signals from compressive measurements via deep learning," International Joint Conference on Neural Networks, pp. 2856-2863, 2016.

[26] http://torch.ch/